\documentclass[conference]{IEEEtran}
\IEEEoverridecommandlockouts
\usepackage{cite}
\usepackage{amsmath,amssymb,amsfonts}
\usepackage{algorithmic}
\usepackage{graphicx}
\usepackage{textcomp}
\usepackage[dvipsnames]{xcolor}
\usepackage{xcolor}
\usepackage{graphicx}

\newcommand{\q}[1]{``#1''}
\usepackage{bm}
\usepackage{multirow}
\usepackage{comment}
\usepackage{pgfplots}
\usepackage{url}

\pgfplotsset{compat=1.18}
\usepgfplotslibrary{fillbetween}
\DeclareGraphicsExtensions{.pdf,.png}

\def\BibTeX{{\rm B\kern-.05em{\sc i\kern-.025em b}\kern-.08em
    T\kern-.1667em\lower.7ex\hbox{E}\kern-.125emX}}
\begin{document}

%\title{Conference Paper Title*\\

\title{Deep multitask neural networks for solving some stochastic optimal control problems}

%\title{Deep multitask neural networks for solving some stochastic optimal control problems\\
%\thanks{The PhD of the first author is funded by Engie Global Markets through CIFRE agreement.}}

\begin{comment}
\author{\IEEEauthorblockN{1\textsuperscript{st} Given Name Surname}
\IEEEauthorblockA{\textit{dept. name of organization (of Aff.)} \\
\textit{name of organization (of Aff.)}\\
City, Country \\
email address or ORCID}
\and
\IEEEauthorblockN{2\textsuperscript{nd} Given Name Surname}
\IEEEauthorblockA{\textit{dept. name of organization (of Aff.)} \\
\textit{name of organization (of Aff.)}\\
City, Country \\
email address or ORCID}
\and
\IEEEauthorblockN{6\textsuperscript{th} Given Name Surname}
\IEEEauthorblockA{\textit{dept. name of organization (of Aff.)} \\
\textit{name of organization (of Aff.)}\\
City, Country \\
email address or ORCID}
}
\end{comment}

%\begin{comment}
\author{\IEEEauthorblockN{Christian Yeo}
\IEEEauthorblockA{\textit{Sorbonne Université, LPSM, UMR 8001, Paris, France} \\
\textit{Engie Global Markets, Courbevoie, France}\\
christian.yeo@sorbonne-universite.fr}
}
%\end{comment}

%\author{\IEEEauthorblockN{Anonymous Authors}}
\maketitle

\begin{abstract}
Most existing neural network-based approaches for solving stochastic optimal control problems using the associated backward dynamic programming principle rely on the ability to simulate the underlying state variables. However, in some problems, this simulation is infeasible, leading to the discretization of state variable space and the need to train one neural network for each data point. This approach becomes computationally inefficient when dealing with large state variable spaces. In this paper, we consider a class of this type of stochastic optimal control problems and introduce an effective solution employing multitask neural networks. To train our multitask neural network, we introduce a novel scheme that dynamically balances the learning across tasks. Through numerical experiments on real-world derivatives pricing problems, we prove that our method outperforms state-of-the-art approaches.
\end{abstract}

\vspace{0.1cm}

\section{Introduction}
\noindent
Multitask learning \cite{b4, b9} is a machine learning field which aims at learning concurrently multiple related tasks at once by leveraging common knowledge among them. This approach employs a unified representation for all tasks, making multiple predictions in a single forward propagation and benefiting from parameters sharing. In contrast, single-task representation entails learning each task independently with a dedicated architecture per task. The multitask learning approach offers advantages such as enhanced data efficiency (particularly valuable in scenarios with limited data whereas most of deep learning-based approaches require huge amount of data) and a reduced risk of overfitting due to shared representation. These features have led to extensive utilization of multitask neural networks across various machine learning domains, including computer vision \cite{b9}, natural language processing (NLP) \cite{b11, b12}, speech recognition \cite{b13, b14}, and reinforcement learning \cite{b15}, among others.   

\vspace{0.1cm}
While multitask learning approaches have been extensively explored in various fields, there still exist unexplored fields. This is the case of stochastic optimal control problems and may be due to two factors. On the one hand, it seems that several learning problems in the aforementioned fields lend themselves quite naturally to a multitask modelling, at least more naturally than in stochastic optimal control problems. On the other hand, classic neural networks have already been successfully applied for solving a wide class of stochastic optimal control problems. However, there exists some class of these problems that classic neural networks fail to solve. Indeed, most of the neural network-based methods require the ability to simulate the underlying state variable which intrinsically assume the knowledge of its distribution. There exist cases where this distribution is unknown, thereby requiring the discretization of the state variable space and the training of one neural network per data point. This becomes numerically impractical when the state variable space is large.

\vspace{0.1cm}
In this paper, we consider a certain class of stochastic optimal control problems which involves the aforementioned issue. We introduce an efficient solution based on multitask learning, enabling us to avoid training one neural network for each data point but rather using a single multitask neural network with shared parameters among tasks. The training of such a multitask neural network involves optimizing multiple loss functions simultaneously, posing a challenge known as \emph{negative transfer}. This issue arises due to dissimilarities among tasks trained together, often forcing the algorithm to optimize a task loss to the detriment of the others. To mitigate negative transfer, various studies have explored what is commonly known as a \emph{loss weighting scheme}. Typically, these schemes aim to adjust task weights during training, increasing the weight for tasks with slow learning speed and vice versa. In some cases \cite{b15, b19, b20, b21}, tasks weights are fixed meaning they are computed explicitly and behave the same throughout the training. In other cases \cite{b5, b22}, weights, additionally to the neural network weights, are considered as learnable parameters. For a comprehensive review of different loss weighting schemes, we refer the reader to \cite{b16, b17, b18}. In this paper, we present a novel loss weighting scheme designed for training multitask neural networks, which belongs to the latter category. At the end of the paper, we perform numerical illustrations of the effectiveness of our multitask approach on real-world pricing problems.

\vspace{0.1cm}
\noindent
\textbf{Contributions.} The main contributions of this paper are:

$\bullet$ We propose a novel loss weighting scheme to train multitask neural networks.

$\bullet$ This paper pioneers the use of neural networks to efficiently solve the dynamic programming principle associated to the class of stochastic optimal control problems discussed below.

$\bullet$ By considering the pricing of a commodity financial product, we numerically prove the effectiveness of our multitask modelling and how it outperforms state-of-the-art methods.

\vspace{0.1cm}
\noindent
\textbf{Related work.} In the multitask learning literature, as mentioned earlier, various loss weighting schemes dynamically increase the weight of a task if it has underperformed in recent iterations. Typically, this is achieved by establishing a \emph{learning speed} metric, often defined as a ratio between the loss at a given iteration and the loss at previous iterations \cite{b24}, initial iteration \cite{b5, b25}, mean of past iteration losses \cite{b26}, and so forth. However, this approach proves impractical in our stochastic optimal control problem, where what we define as \q{loss} function may be null or even negative, rendering certain loss weighting schemes like \emph{GradNorm} \cite{b5} ineffective. Besides, when addressing the class of stochastic optimal control problems considered in this paper, to the best of our knowledge, there are few, if any, studies which consider the resolution of the associated dynamic programming principle with neural networks. This is primarily due to the inefficiency of most neural network-based methods in handling such problems. In a previous work \cite{b23}, the authors proposed various numerical methods with an application to the pricing of \emph{storage contracts} which fall within the framework considered in this paper. Due to the lack of knowledge about the distribution of the underlying state variables, the authors in \cite{b23} used a proxy by drawing from a uniform distribution. In this paper, we prove that, in such problems, we can overcome the need for such approximations through the use of a well-designed multitask modelling. The code for the numerical experiments is available at: \textcolor{magenta}{\url{https://anonymous.4open.science/r/MultiTask_Swing-AE0B}}.

\section{Stochastic optimal control and dynamic programming}
\noindent
We set a time grid $\{0 = t_0 < t_1 < \ldots < t_n = T\}$ and consider a state process $(s_k)_{0 \le k \le n}$ which is a $\mathbb{R}^d$-valued discrete Markov process with $s_0$ assumed to be deterministic. We set a filtered probability space $\mathcal{P} := \big(\Omega, \mathcal{F}, (\mathcal{F}_{k})_{0 \le k \le n}, \mathbb{P}\big)$, where $\mathcal{F}_{k}$ is the natural completed filtration of the random vector $s_k$. Unless otherwise stated, any variable indexed by $k$ will be assumed to be $\mathcal{F}_k$-measurable.

\subsection{Stochastic optimal control problem}
\noindent
We consider the following class of Stochastic Optimal Control (\emph{SOC}) problems on $\mathcal{P}$:
\begin{equation}
\small
    \label{soc_pb}
    P_0 := \underset{(q_{k})_{0 \le k \le n-1} \in \mathcal{U}}{\sup} \hspace{0.1cm} \mathbb{E}\Bigg[\sum_{k=0}^{n-1} c_{k}\big(s_{k}, q_{k}(s_k)\big) + g_n(s_n, Q_n) \Bigg],
\end{equation}
where the function $c_k : \mathbb{R}^d \times \mathbb{R}_{+} \to \mathbb{R}$ is the immediate reward function at time $t_k$ and the controls space $\mathcal{U}$ is defined by
\begin{equation}
\footnotesize
    \label{set_adm_strat}
    \mathcal{U} := \Big\{(q_{k})_{0 \le k \le n-1}, \quad \underline{q} \le q_{k} \le \overline{q} \quad \text{and}  \quad \underline{Q} \le Q_n:=\sum_{k = 0}^{n-1} q_{k} \le \overline{Q}  \Big\}.
\end{equation}
\noindent
Parameters $0 \le \underline{q} < \overline{q}$ and $\underline{Q} \le \overline{Q}$ are integers. The terminal condition $g_n : \mathbb{R}^d \times \mathbb{R}_{+} \to \mathbb{R}$ is a function depending on the state variable $s_n$ at date $t_n$ and the cumulative control up to date $t_{n-1}$. In the sequel, we define the cumulative control at a date $t_k$ as following:
\begin{equation}
    \label{def_cum_control}
    Q_k = \sum_{\ell = 0}^{k-1} q_\ell \quad \text{with} \quad Q_0 =0.
\end{equation}
\noindent
At each date $t_k$, due to the double constraint defining the preceding set $\mathcal{U}$, we must have
\begin{equation}
    \label{interval_Q}
    Q_k \in \mathcal{Q}_k := \big[Q_k^d, Q_k^u \big]
\end{equation}
with
\begin{equation}
    \label{def_var_interval_Q}
    Q_k^d := \max\big(k \underline{q}, \underline{Q} - (n-k) \overline{q}\big) \quad Q_k^u := \min\big(k \overline{q}, \overline{Q}\big).
\end{equation}

\begin{figure}[htbp]
    \centering
    \includegraphics[width=1.0\columnwidth]{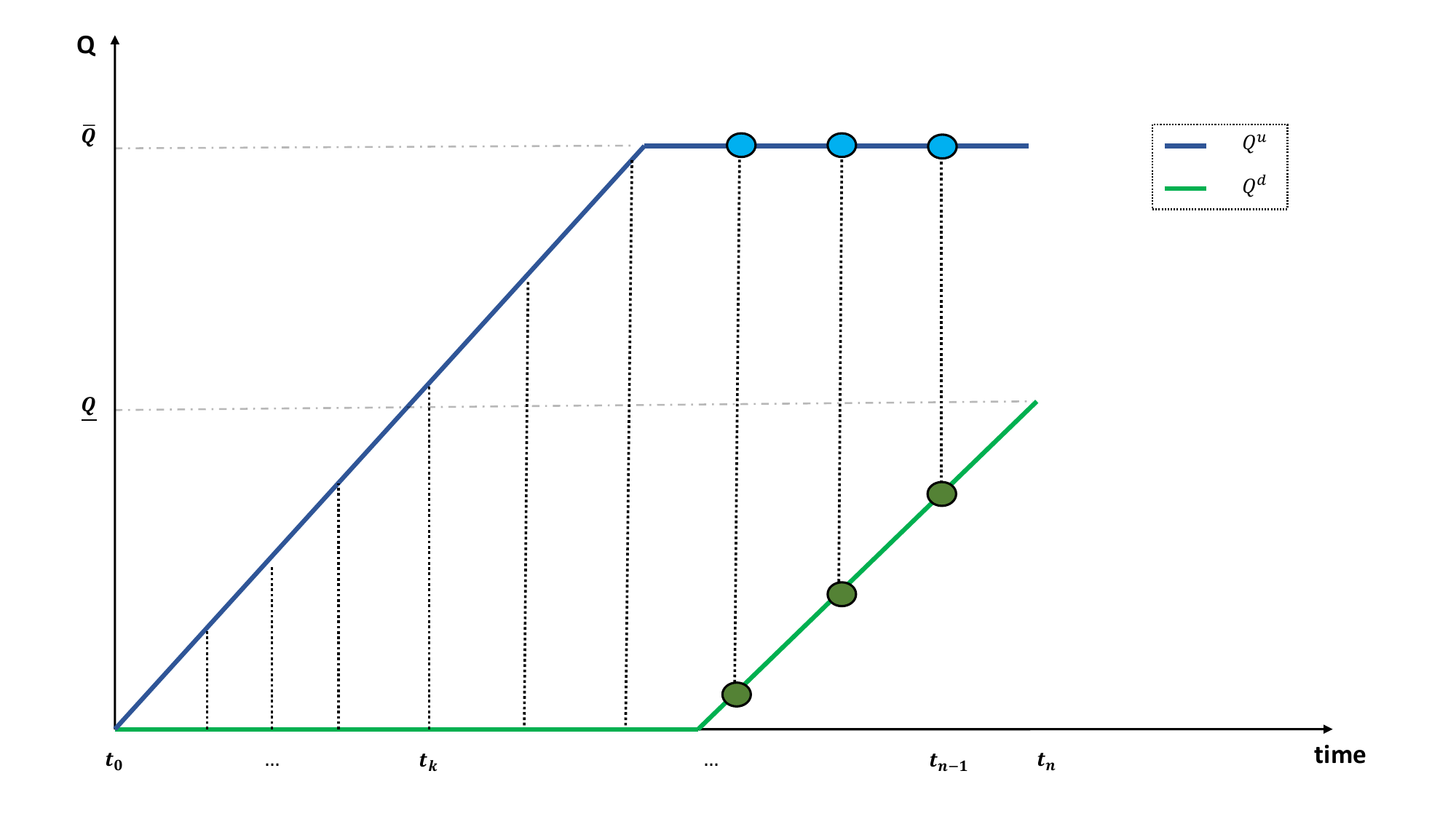}
    \caption{Illustration of the space of cumulative controls. We used $\underline{q} = 0$. Blue and green points will be called \emph{trivial tasks} in the subsequent analysis.}
    \label{volume_grid}
\end{figure}

\noindent
This allows to draw the cumulative controls space (see Fig. \ref{volume_grid}) which represents attainable cumulative controls at each date.

\vspace{0.1cm}
The \emph{SOC} problem \eqref{soc_pb}, where the sequence $(q_k)_{0 \le k \le n-1}$ represents the controls at each date, can be solved by the \emph{Backward Dynamic Programming Principle (BDPP)}. Moreover, this \emph{SOC} problem includes the modelling of several commodity derivative products such as \emph{swing} \cite{b1,b2,b3} and \emph{storage} contracts \cite{b23,b3}.

\subsection{Backward dynamic programming principle}
\noindent
To state the \emph{BDPP}, we start by introducing \eqref{soc_pb} seen from date $t_k$. That is, for a $\mathcal{F}_{k-1}$-measurable random variable, $Q_k \in \mathcal{Q}_k$, we consider
\begin{equation}
\footnotesize
    \label{swing_price_interm}
    V_k(s_k, Q_k) := \underset{(q_{\ell})_{k \le \ell \le n-1} \in \mathcal{U}_k}{\sup} \hspace{0.1cm} \mathbb{E}\Bigg[\sum_{\ell=k}^{n-1} c_{\ell}\big(s_{\ell}, q_{\ell}(s_\ell)\big) + g_n(s_n, Q_n) \Big\rvert \mathcal{F}_k \Bigg],
\end{equation}
\noindent
where the controls space $\mathcal{U}_k$ from date $t_k$ is defined by
\begin{equation}
    \label{set_U_k}
    \mathcal{U}_k := \Big\{(q_{\ell})_{k \le \ell \le n}, \quad (\underline{Q} - Q_k)_{+} \le \sum_{\ell = k}^{n-1} q_{\ell} \le \overline{Q} - Q_k  \Big\}.
\end{equation}
\noindent
Note that, one has $V_0(s_0, 0) = P_0$ (recalling that $s_0$ is deterministic). Besides, one may show \cite{b1, b2, b3} that, for each date $t_k$, there exists a \emph{value function} $\mathbb{R}^d \times \mathcal{Q}_k \ni (s, Q) \mapsto v_k(s, Q)$ such that, almost surely, $v_k(s_k, Q) = V_k(s_k, Q)$ and which is defined by the following dynamic programming equation
\begin{equation}
\label{bdpp}
\small
\left\{
    \begin{array}{ll}
        v_{n}(s, Q) = g_n(s, Q),\\
        v_k(s, Q) = \underset{q \in \Gamma_k(Q)}{\sup} \Big[c_k(s, q) + \mathbb{E}\big(v_{k+1}(s_{k+1}, Q+q) \rvert s_k = s \big)\Big],
    \end{array}
\right.
\end{equation}
\noindent
where, for each date $t_k$ and any $Q \in \mathcal{Q}_k$, the set $\Gamma_k(Q)$ of admissible controls at that date is defined by the interval:
\begin{equation}
    \label{set_adm_control}
    q_k(Q) \in \Gamma_k(Q) := \big[q_k^{-}(Q),  q_k^{+}(Q)\big]
\end{equation}
\noindent
with
\begin{equation*}
    \label{def_borne_interval_adm_q}
    q_k^{-}(Q) := \max\big(Q_{k+1}^d - Q, \underline{q}\big) \quad q_k^{+}(Q) := \min\big(Q_{k+1}^u - Q, \overline{q} \big).
\end{equation*}

\vspace{0.1cm}
\noindent
\textbf{Bang bang control.} 
We will assume that the range $\overline{Q} - \underline{Q}$ is a multiple of $\overline{q} - \underline{q}$ so that the set $\mathcal{Q}_k$ is made of integers i.e.
\begin{equation}
\label{interval_Q_discret}
\mathcal{Q}_k = \mathbb{N} \cap \big[Q_k^d, Q_k^u \big].
\end{equation}
In this context, $I_k$ is defined as the cardinality of the finite set $\mathcal{Q}_k$, and we will frequently use the notation $\mathcal{Q}_k = \{Q_k^1,\ldots,Q_k^{I_k}\}$. Furthermore, it has been established in \cite{b2,b3} that this condition implies that the optimal control is \emph{bang-bang}, meaning it can only take two values i.e., for any date $t_k$ and $Q \in \mathcal{Q}_k$, $q_k(Q) \in \{q_k^{-}(Q), q_k^{+}(Q)\}$. In the subsequent analysis, we operate within this bang-bang framework.

\vspace{0.2cm}
\noindent
\textbf{Problematic.} 
The joint distribution of the state variable $(s, Q) \in \mathbb{R}^d \times \mathbb{R}_{+}$ is not known. Consequently, practically solving \eqref{bdpp} involves computing, for each date $t_k$ and each potential cumulative control $Q \in \mathcal{Q}_k$, the optimal control $q_k(Q)$. This optimal control solves \eqref{bdpp} at that specific date, maximizing the expression $q \mapsto c_k(s, q) + \mathbb{E}\big(v_{k+1}(s_{k+1}, Q+q) \rvert s_k = s \big)$. A straightforward application of neural networks would be inefficient because it would require training a separate neural network for each date $t_k$ and each possible cumulative control $Q \in \mathcal{Q}_k$. This approach may become numerically impractical, especially when the set $\mathcal{Q}_k$ is large.

\vspace{0.1cm}
We propose an efficient alternative based on \emph{multitask neural networks} \cite{b4}, enabling the training of a single neural network per date $t_k$. Specifically, at each date $t_k$, our neural network is designed as a multitask neural network, capable of simultaneously computing all optimal controls $q_k(Q_k^1),\ldots,q_k(Q_k^{I_k})$ in one forward propagation. The architecture of our multitask neural network and its training procedure are detailed in the subsequent section.

\section{Multitask learning}
\noindent
We initially express our \emph{SOC} problem \eqref{soc_pb} as a multitask learning problem through the BDPP \eqref{bdpp}, and propose its solution using multitask neural networks. Subsequently, we present one of the main contribution of this paper concerning the training of the multitask neural network by introducing a scheme that dynamically balances the learning between tasks that need to be trained together.

\subsection{Towards multitask learning}
\noindent
For each date $t_k$, the optimal control is \emph{bang-bang} so that the supremum in \eqref{bdpp} is a maximum between two values. Therefore, for $k=0,\ldots,n-1$ and $i=1,\ldots,I_k$, there exists a Borel set $A_k^i \in \mathcal{F}_k$ such that almost surely:
\begin{equation}
    \label{bang_bang_v}
    v_k(s_k, Q_k^i) = \mathbb{E}\Big[\psi_k^{+}(Q_k^i) \mathbf{1}_{A_{k}^i} + \psi_k^{-}(Q_k^i) \big(1- \mathbf{1}_{A_{k}^i}\big) \Big\rvert s_k \Big],
\end{equation}
\noindent
where,
\begin{equation}
\label{payoff_bang_bang}
\psi_k^{\pm}(Q) := c_k\big(s_k, q_k^{\pm}(Q)\big) + v_{k+1}\big(s_{k+1}, Q+q_k^{\pm}(Q)\big).
\end{equation}
\noindent
Note that, as we proceed backwardly, when we compute $\psi_k^{\pm}$ at date $t_k$, the value function $v_{k+1}$ is known. 

\vspace{0.2cm}
Next, our objective is to simultaneously approximate all \emph{decision functions}, denoted by indicator functions $\big(\mathbf{1}_{A^i_{k}}\big)_{1 \le i \le I_k} \in \{0,1\}^{I_k}$, using a single parametric function with values in $[0, 1]^{I_k}$. This chosen parametric function is a multitask neural network. Throughout this paper, at a given date $t_k$, we will use the term \q{\emph{task}} to signify the approximation of one decision function $\mathbf{1}_{A^i_{k}}$ ($i = 1,\ldots,I_k$). Thus, at each date $t_k$, we have $I_k$ tasks.

\begin{figure}[htbp]
    \centering
    \includegraphics[width=1.0\columnwidth]{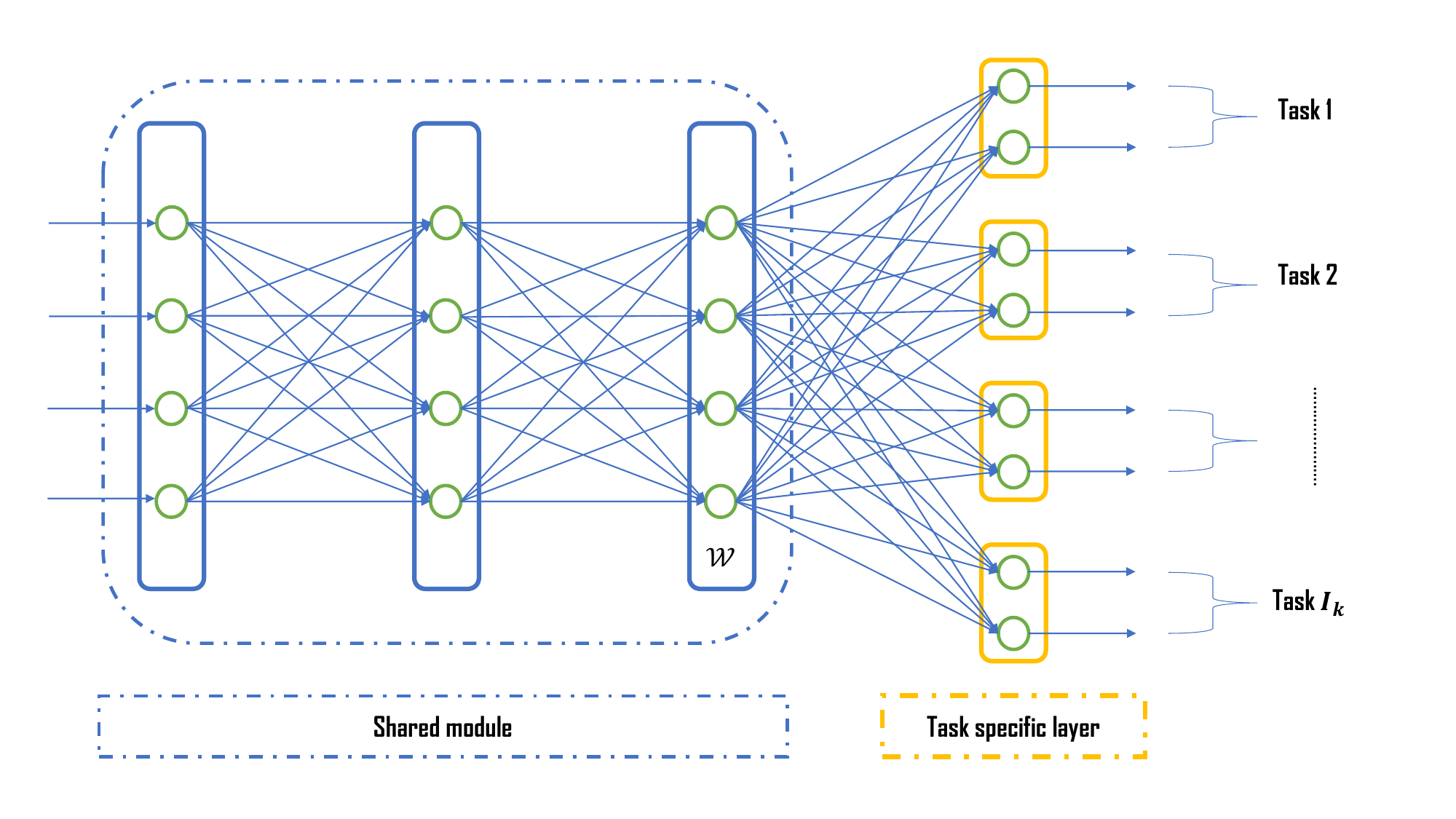}
    \caption{Multitask network architecture. The shared module is made of the input layer and 2 hidden layers. There are $I_k=4$ task specific layers for each task. Each output $\chi_k^{(i)}(\cdot;\theta)$ ($i=1,\ldots,I_k$) of task specific layers is a $\mathbb{R}^2$-valued vector used to define our approximation of the decision function as described in \eqref{def_decision_function_f}.}
    \label{mtl_representation}
\end{figure}

\vspace{0.1cm}
Our modelling, based on \emph{hard parameter sharing} architecture \cite{b4}, is the following. We first take into account the fact that, as claimed in \cite{b7}, in this type of \emph{SOC} problems, the \emph{remaining capacity} $m(Q) := (Q-\underline{Q})/(\overline{Q}-\underline{Q})$, is an important feature to help finding the optimal control. Then, we build a multitask neural network defined as follows
\begin{equation}
    \label{mtl_def_chi}
    \chi_k(\cdot;\theta_k) : \mathbb{R}^d \to \mathcal{M}(2 \times I_k),
\end{equation}
\noindent
where $\mathcal{M}(2 \times I_k)$ denotes the set of $2 \times I_k$ real matrix. In other words, our multitask neural network has $I_k$ $\mathbb{R}^2$-valued outputs. Besides, it is made of two components (see Fig. \ref{mtl_representation}). A \emph{shared module} made of a sequence of hidden layers which are common to all tasks. This allows to leverage knowledge among tasks to make a generalized feature thus reducing the risk of overfitting. At the head of the model are \emph{task specific layers} which are, as their name suggest, specific layers to each task.

\vspace{0.1cm}
Let $i \in \{1,\ldots,I_k\}$. The $\mathbb{R}^2$-valued $i^{th}$ output of the multitask network \eqref{mtl_def_chi}, denoted by $\chi_k^{(i)}$, is then composed with the Sigmoid function $\sigma(x) := 1 /(1+e^{-x})$ to map $\mathbb{R}$ to $[0,1]$. That is, we define our approximation for the decision function $\mathbf{1}_{A^i_k}(\cdot)$ by:
\begin{equation}
    \label{def_decision_function_f}
    f_k^{(i)}(s_k; \theta_k) := \sigma\big(\langle \chi_k^{(i)}(s_k; \theta_k), Z_k^i \rangle \big),
\end{equation}
where $\langle\cdot,\cdot \rangle$ denotes the Euclidean inner-product and $Z_k^i = (M(Q_k^i), 1)^\top \in \mathbb{R}^2$. Note that the Sigmoid function $\sigma$ is chosen because it is a $[0,1]$-valued and sufficiently smooth function to facilitate \cite{b27} the application of automatic differentiation algorithms in backpropagation.

\vspace{0.1cm}
It is important to observe that in \eqref{def_decision_function_f}, for the sake of simplicity in notations, we have opted to express the $i^{th}$ approximation as if it were dependent on the parameters $\theta_k$ of the entire multitask network. In reality, this approximation relies on a reduced set of parameters, namely the shared module parameters along with the parameters of the $i^{th}$ task-specific layer. This choice of notation does not impact the analysis.

\vspace{0.1cm}
The parameters $\theta_k$ are then found such that they maximize, at once and in average, all random variables in \eqref{bang_bang_v} for all $i=1,\ldots,I_k$. More precisely, owing to the \emph{tower rule} applied to \eqref{bang_bang_v} and replacing the decision function by our multitask network approximation, at each date $t_k$, we want to minimize, at once and with the same parameters $\theta_k$, all \emph{loss functions} ($i=1,\ldots,I_k$)
\begin{equation}
\small
    \label{def_loss_per_task}
    \mathcal{L}_{k, i}(\theta_k) := -\mathbb{E}\Big[\psi_k^{+}(Q_k^i) f_k^{(i)}(s_k; \theta_k) + \psi_k^{-}(Q_k^i) \big(1- f_k^{(i)}(s_k; \theta_k)\big)\Big]
\end{equation}

\noindent
which is the aim the following section. Besides, notice that, to minimize these loss functions, we rely on \emph{Stochastic Gradient Descent}. The differentiation is performed using existing libraries for \emph{Automatic Algorithm Differentiation} with \emph{Adam} optimizer \cite{b6}. For the purpose of a comprehensive analysis, in the subsequent section, we omit the time index $k,$ as the same operation will be conducted at each time step.

\subsection{Training multitask neural network}
\noindent
Our objective is to minimize multiple loss functions at once and possibly concurrently. To this end, we consider a \emph{multitask learning} framework \cite{b4}. Let us recall the background.

\vspace{0.1cm}
\noindent
\textbf{Training phase.} We consider $I$ learning tasks with corresponding loss functions $\mathcal{L}_i^{(t)}$ ($i=1,\ldots,I$) at iteration $t$. We want to minimize each task's loss function $\mathcal{L}_i$ at once with respect to the same set of parameters. To this end, we define a \emph{global loss function} as a single aggregated version of individual loss functions i.e.
\begin{equation}
    \label{global_loss}
    \mathcal{L}^{(t)} := \sum_{i = 1}^I w_i^{(t)} \cdot \mathcal{L}_i^{(t)},
\end{equation}
\noindent
where $w_i^{(t)}$ is the weight of task $i$ at iteration $t$. In multitask learning literature, this way of defining the global loss function is known as \emph{Linear Scalarization} and is the most commonly-used approach for solving multitask learning problems.

\vspace{0.1cm}
The training process of a multitask network involves the simultaneous minimization of all loss functions $\mathcal{L}_i$ only based on the minimization of the global loss function $\mathcal{L}$. This task is recognized as challenging due to the presence of \emph{negative transfer}. The latter occurs when certain tasks dominate the training process, negatively affecting the performance of others. This underscores the importance of selecting \q{optimal} weights $(w_i)_{1 \le i \le I}$ to achieve a proper balance in learning between tasks. Essentially, this involves adjusting the task's loss weight in such a way that it is increased when the \q{\emph{learning speed}} for that task is low. This has to be done so that parameters of the multitask network converge to robust shared features that prove beneficial across all tasks. In this paper, we propose a dynamic weight scheme that automatically (1) normalizes task losses to ensure each task is trained at a similar rate, and (2) balances learning across tasks.

\vspace{0.1cm}
\noindent
\textbf{Sigmoid-Moving Average GradNorm (\emph{S-MAG}).}
We build upon \emph{GradNorm} scheme \cite{b5}. This dynamic weight scheme may suffer from two issues. Firstly, like several weight schemes, it may not directly address learning problems where the \q{loss} function can be null and potentially negative (as is the case for our loss functions \eqref{def_loss_per_task} in the opposite of most, if not all, learning problems where the loss function is nonnegative). Additionally, it may exhibit a strong dependence on the initialization process, as the aforementioned learning speed is based on how the loss at an iteration deviates from the initial loss. To address these concerns, we introduce the \emph{Sigmoid-Moving Average GradNorm (S-MAG)} scheme.

\vspace{0.1cm}
To present \emph{S-MAG} scheme, let us start by defining the same quantities as in GradNorm. That is, at iteration $t$, we consider:

\vspace{0.1cm}
\noindent
$\bullet$ $\bm{\mathcal{W}}$: network weights of the last shared layer (as indicated in Fig. \ref{mtl_representation}).

\vspace{0.2cm}
\noindent
$\bullet$ $\bm{G_{\mathcal{W}}^{(t)}(i)} := \big|\big|\nabla_{\mathcal{W}} \hspace{0.1cm} w_i^{(t)} \cdot \mathcal{L}_i^{(t)} \big|\big|_2$: $L2$-norm of the gradient of the weighted single task loss.

\vspace{0.2cm}
\noindent
$\bullet$ $\bm{\bar{G}_{\mathcal{W}}^{(t)}} :=  \frac{1}{I} \sum_{i = 1}^I G_{\mathcal{W}}^{(t)}(i)$: average gradient norm across all tasks.

\vspace{0.2cm}
Then, for all tasks $i =1,\ldots,I$, we introduce the following \emph{learning speed} quantities:

\vspace{0.2cm}
\noindent
$\bullet$ $\bm{\Tilde{L}^{(t)}(i)} := \sigma\big(\mathcal{L}_i^{(t)} - \bar{\mathcal{L}}_i^{(t)} \big)$, where $\sigma$ is the Sigmoid function. $\bar{\mathcal{L}}_i^{(t)}$ denotes the exponential moving average of past loss values:
\begin{equation}
\label{ema_loss_def}
\bar{\mathcal{L}}_i^{(0)} = \mathcal{L}_i^{(0)} \quad \text{and} \quad \bar{\mathcal{L}}_i^{(t)} = \beta \bar{\mathcal{L}}_i^{(t-1)} + (1-\beta)\mathcal{L}_i^{(t-1)},
\end{equation}
\noindent
where $\beta \in [0,1]$ is the decay rate of the exponential moving average. Therefore, $\Tilde{L}^{(t)}(i)$ serves as a representation of the \emph{inverse training speed}. Indeed, note that, given our minimization problem, when $\mathcal{L}_i^{(t)}$ exceeds $\bar{\mathcal{L}}_i^{(t)}$, task $i$ exhibits a poor learning speed trend, resulting in a higher value for the quantity $\Tilde{L}^{(t)}(i)$. Besides, as mentioned earlier, this framework accommodates possible null loss functions and offers the benefit of smoothing the estimation of losses used as benchmarks to define the learning speed.

\vspace{0.2cm}
\noindent
$\bullet$ $\bm{r^{(t)}(i)} := \frac{\Tilde{L}^{(t)}(i)}{\frac{1}{I} \sum_{i = 1}^{I} \Tilde{L}^{(t)}(i)}$: the relative inverse learning speed. A higher value of this quantity indicates that the $i^{th}$ task performed poorly compared to other tasks, possibly due to being dominated.

\vspace{0.2cm}
We then treat weights $w_i^{(t)}$, for all $i=1,\ldots,I$, as a learnable parameter and we update them using stochastic gradient descent with the aim of solving the following optimization:
\begin{equation}
    \label{gradnorm_loss}
    \underset{w_1, \ldots, w_I}{\min} \sum_{i = 1}^{I} \Big|G_{\mathcal{W}}^{(t)}(i) -  \bar{G}_{\mathcal{W}}^{(t)} \times \big[r^{(t)}(i)\big]^{\alpha} \Big|_1,
\end{equation}
\noindent
where $\alpha \ge 0$ is a hyperparameter and $|\cdot|_1$ denotes the Euclidean $L1$-norm. Setting the task's weight through the solution of \eqref{gradnorm_loss} dynamically increases the gradient for tasks with poor learning speed trends. The hyperparameter $\alpha$ governs this increase. A low $\alpha$ results in tasks being trained at nearly the same rate, suitable for very similar tasks. Conversely, a high $\alpha$ is chosen to enforce balance between dissimilar tasks. Besides it is important to note that, when differentiating \eqref{gradnorm_loss}, the quantity $\bar{G}_{\mathcal{W}}^{(t)} \times \big[r^{(t)}(i)\big]^{\alpha}$ is held constant. Additionally, similar to GradNorm, we renormalize the weights $w_i^{(t)}$ after each iteration, ensuring that $\sum_{i=1}^{I} w_i^{(t)} = I$, to separate gradient normalization from the global learning rate. At the first iteration, we set $w_i^{(0)} = 1$ for all $i=1,\ldots,I$.

\vspace{0.1cm}
To conclude this section, it is important to highlight that the dynamic loss weighting scheme (S-MAG scheme) introduced here can be applied to learning problems beyond our SOC problem \eqref{soc_pb}. We assert that it can be tested for all types of problems covered by multitask learning theory, such as those encountered in Computer Vision.

\section{Experiments}
\label{sec_exp}
\noindent
We numerically prove the effectiveness of our multitask modelling. To this end, we consider the pricing of two derivative products namely, the \emph{Take-or-Pay} (\emph{ToP}) contract and the swing contract with penalty. For both cases, we compare our multitask network to state-of-the-art (\emph{SOTA}) methods.
\subsection{Implementation details}
\label{sous_sec_exp_set}

\noindent
This section aims at providing some implementation details complementing the methodology described above.

\begin{figure}[httb]
    \centering
    \includegraphics[width=1.0\columnwidth]{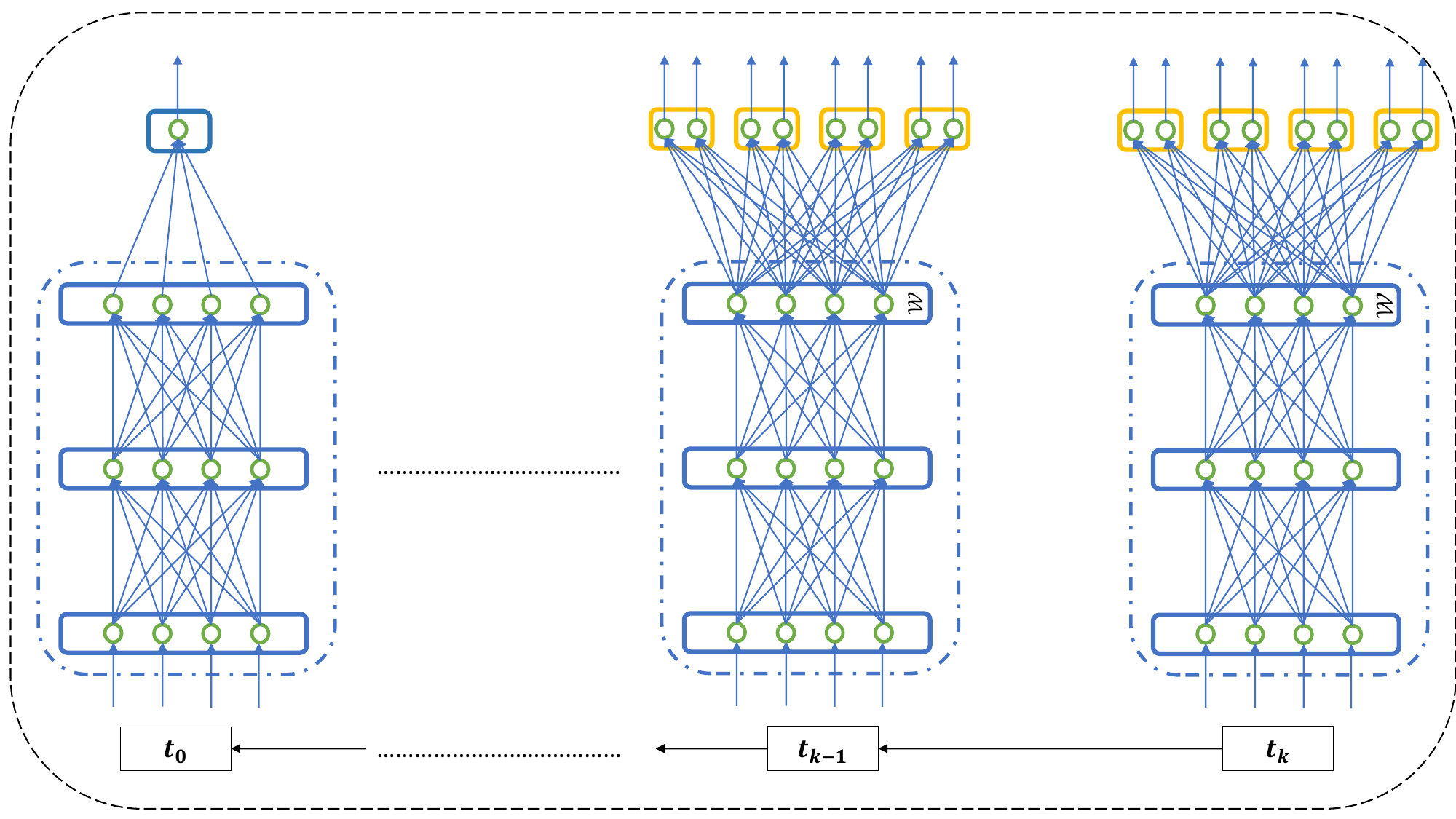}
    \caption{Illustration of the deep backward multitask network. One $\mathbb{R}$-valued feedforward neural network for date $t_0$ and the remaining are multitask neural networks.}
    \label{global_model}
\end{figure}

\vspace{0.2cm}
\noindent
\textbf{Detail I.} At date $t_0 = 0$, we have $I_0 = 1$ since the set $\mathcal{Q}_0$ reduces to the singleton $\{Q_0 = 0\}$. Whence, there is just one task to train. Therefore, practically, we consider a list of $n$ (deep) neural networks, where the first one is a classic $\mathbb{R}$-valued feedforward neural network with the Sigmoid function as activation function for the output layer. The others $n-1$ are multitask networks (see also Fig. \ref{global_model}). 

\vspace{0.1cm}
\noindent
\textbf{Detail II.} As previously stated, when addressing the BDPP \eqref{bdpp} at date $t_k$, the value function $v_{k+1}$ is predetermined. In practice, it is considered in its \emph{bang-bang} form. The latter involves substituting the approximating decision function $f_{k+1}^{(i)}(\cdot; \theta) \in [0,1]$ with its \q{bang-bang version} $F_{k+1}^{(i)}(\cdot; \theta)$, where
\begin{equation}
    \label{bang_bang_f}
    F_{k+1}^{(i)}(s_{k+1};\theta_{k+1}) := \mathbf{1}_{\{f_{k+1}^{(i)}(s_{k+1}; \theta_{k+1}) \ge 1/2\}} \in \{0,1\}.
\end{equation}

\vspace{0.1cm}
\noindent
\textbf{Detail III: Remove trivial tasks.} At each date, we remove, if any, \emph{trivial tasks}. The latter denotes tasks that are highlighted (blue and green circles) in Fig. \ref{volume_grid}. Indeed, for these tasks, due to constraints space, there is only one feasible option between $q_k^{-}, q_k^{+}$. For blue circles, only $q_{k}^{-}$ is feasible, and for green circles, only $q_k^{+}$ is feasible.

\vspace{0.1cm}
\noindent
\textbf{Detail IV: Valuation phase.} Once our multitask network is trained, we need to perform a \emph{valuation phase} in a forward way in order to compute an estimator of $P_0$ \eqref{soc_pb}. That is, we simulate $M$ \emph{i.i.d.} copies $\big(s_k^{[m]}\big)_{0 \le k \le n, 1 \le m \le M}$ of $(s_k)_{0 \le k \le n}$ and consider the estimator:
\begin{equation}
\small
    \label{swing_price_forward_val}
     \widehat{P_0} := \frac{1}{M} \sum_{m = 1}^M \sum_{k=0}^{n-1} c_k\Big(s_k^{[m]}, q_k^{*, [m]}\big(Q_k^{[m]}\big)  \Big) + g_n\big(s_n^{[m]}, Q_n^{[m]} \big)
\end{equation}
where for $Q_k^i \in \mathcal{Q}_k$, ($i =1,\ldots, I_k$), the optimal control on each path $m$ is defined by
\begin{equation}
\small
    \label{q_star}
    q_k^{*, [m]}(Q_k^i)  := q_k^{-}(Q_k^i) + \big(q_k^{+}(Q_k^i)-q_k^{-}(Q_k^i) \big) \cdot  F_k^{(i)}(s_k^{[m]}; \theta_k^{*})
\end{equation}
\noindent
with $(\theta_k^{*})_{0 \le k \le n}$ being parameters obtained in the training phase. The cumulative control along path $m$ is also updated in a forward way as follows
\begin{equation}
    \label{update_Q}
    Q_0^{[m]} = 0 \quad \text{and} \quad Q_{k+1}^{[m]} = \sum_{\ell = 0}^{k} q_\ell^{*, [m]}(Q_\ell^{[m]}).
\end{equation}
\noindent

\vspace{0.1cm}
\noindent
\textbf{Detail V: Transfer learning.} To speed up the training, we use transfer learning across dates. That is, at a date $t_k$, the shared module parameters are initialized with the just trained shared module parameters at date $t_{k+1}$. Our global model comprising one feedforward neural network and $n-1$ multitask neural networks is depicted in Fig. \ref{global_model}.

\vspace{0.2cm}
Using \emph{PyTorch} toolbox, we implemented our neural networks as follows. For the first date $t_0 = 0$, we use a classic feedforward neural network with one hidden layer and 50 units. For each of the remaining dates, we use a multitask neural network where the shared module is composed of 2 hidden layers with 50 units per layer. In all cases, a batch normalization is used as well as the \emph{ReLU} function (i.e. $x \mapsto \max(x, 0)$) as activation function. For the training phase, we used 200 iterations, the batch size is set to 2048 and the learning rates are $0.1$ and $0.01$ to update respectively the network weights and the task's weights. For the valuation phase, the sample size $M$ is set to $2\cdot10^6$. For \emph{S-MAG} scheme, we set the hyperparameter $\alpha=1.8$ and $\beta =0.7$.

\vspace{0.1cm}
As already mentioned, we consider the pricing of \emph{Take-or-Pay} contract and swing contract with penalty. Both are two types of swing contracts. A swing contract is a commodity derivative product which gives the right to its holder to purchase amounts of energy (for example gas) $q_k$ at dates $t_k$ ($k=0,\ldots,n-1$) and at a predetermined price $K$ often called \emph{strike price}. We suppose that the underlying energy product is a forward contract which price at date $t$ is denoted $F_{t, T}$ and follows the following dynamics,
\begin{equation}
\label{log_normal_dyn}
	\frac{dF_{t, T}}{F_{t_, T}} = \sum_{i = 1}^{d}  \sigma_i e^{-\alpha_i (T-t)}dW_t^i, \quad t \le T,
\end{equation}
\noindent
where for all $1 \le i \le d$, $\big(W_t^i\big)_{t\ge 0}$ is either a Brownian motion (in case $d=1$) or are correlated Brownian motions (in case $d \ge 2$) with the instantaneous correlation given by,
\begin{equation*}
\langle dW_{\cdot}^i,dW_{\cdot}^j \rangle_t =  \left\{
    \begin{array}{ll}
        dt \hspace{1.4cm} \text{if} \hspace{0.2cm} i = j\\
        \rho_{i, j} \cdot dt \hspace{0.65cm} \text{if} \hspace{0.2cm} i \neq j
    \end{array}
\right.
\end{equation*}

\noindent
where $\langle \cdot, \cdot\rangle_t$ denotes the quadratic variation at time $t$. In the diffusion model \eqref{log_normal_dyn}, the spot price is given by a straightforward application of Itô formula:
\begin{equation}
\label{spot_price}
S_k := F_{t_k, t_k} = F_{0,t_k} \cdot \exp\Big(\langle \sigma, s_k \rangle - \frac{1}{2}\lambda_k^2\Big),
\end{equation}
\noindent
where $\sigma = \big(\sigma_1, \ldots, \sigma_d)^\top$, $s_k = \big(s_k^1, \ldots, s_k^d \big)^\top $ and for all $1 \le i \le d$,
\begin{equation*}
s_k^i = \int_{0}^{t_k} e^{-\alpha_i (t_k-s)}\, \mathrm{d}W_s^i \hspace{0.1cm}
\end{equation*}
\noindent
and
\begin{equation*}
    \lambda_k^2 = \sum_{i = 1}^{d} \sum_{j =1}^d  \rho_{i, j} \frac{\sigma_i \sigma_j}{\alpha_i + \alpha_j} \Big(1 - e^{-(\alpha_i + \alpha_j)t_k}  \Big).
\end{equation*}
\noindent
Note that in case $d=1$, coefficient $\lambda_k^2$ reads
\begin{equation*}
    \lambda_k^2 := \frac{\sigma^2}{2 \alpha}\big(1-e^{-2\alpha t_k}\big).
\end{equation*}

\vspace{0.2cm}
The holder of a swing contract is allowed to purchase, at each date $t_k$, an amount $q_k \in [\underline{q}, \overline{q}]$ (\emph{local constraints}) such that at the expiry we have $\sum_{k = 0}^{n-1} q_k \in [\underline{Q}, \overline{Q}]$ (\emph{global constraints}). The \emph{Take-or-Pay} contract corresponds to the swing contract where the latter volume constraints are firm. That is, the holder of the contract is obliged to respect both local and global constraints. In the swing contract with penalty, the holder of the contract may violate global constraints but is faced to a penalty. In the sequel, we will denote by $S$ the spot price as function of the state variable $s$ (as in our model \eqref{spot_price}).

\subsection{Take-or-Pay contract}
\noindent
We first consider a Take-or-Pay (\emph{ToP}) contract \cite{b10}. Here, the holder of the contract wants to find a purchase strategy that solves the \emph{SOC} problem \eqref{soc_pb} where (we assume there are no interest rates),
\begin{equation}
    \label{def_soc_pb_esp_swing}
    g_n \equiv 0 \quad \text{and} \quad c_k(s, q) := q \cdot (S - K).
\end{equation}

We start by proving the effectiveness of our multitask modelling. For this purpose, we compare our approach with two benchmarks. The first benchmark is a deep neural network-based method referred to as \emph{NN strat} \cite{b7}. The second benchmark is the Longstaff and Schwartz method \cite{b3}, which solves the BDPP by replacing the conditional expectation with an orthogonal projection onto a subspace spanned by a finite number of functions.

\vspace{0.2cm}
\noindent
\textbf{Benchmark setting.} We set $\underline{q} = 0, \overline{q} = 6, \overline{Q} = 200, F_{0, t_k} = 20, K = 20$. Results are recorded in Table \ref{table_results_benchmark}. In the remaining, we will consider basic weights strategies namely, \emph{EW} (\emph{equal weights}) and \emph{UW} (\emph{uniform weights}) which denote respectively the case where weights are equal and the one where they are drawn from $\mathcal{U}(0,1)$. It is worth mentioning that we chose not to implement the single-task approach as it requires significant resources.

\begin{table}[htbp]
\caption{Comparison of the multitask neural network with state-of-the-art methods for the pricing of Take-or-Pay contracts and considering a benchmark example. All methods are based on the dynamic programming principle except \emph{NN-strat} which is based on a global optimization approach.}
\begin{center}
\begin{tabular}{|c|c|c|c|c|c|}
\hline
& \multicolumn{3}{c|}{Multitask} & \multicolumn{2}{c|}{Benchmark}\\
\hline
& \textbf{S-MAG}& \textbf{EW}& \textbf{UW} & \textbf{NN-strat}& \textbf{LS} \\
\hline
ToP ($\underline{Q} = 140)$ & \textbf{65.44} & 65.13 & 65.14 & 65.27 & 65.14 \\
ToP ($\underline{Q} = 170)$ & \textbf{23.91} & 23.57 & 23.57 & 23.82 & 23.72 \\
\hline
\end{tabular}
\label{table_results_benchmark}
\end{center}
\end{table}

\vspace{0.1cm}
Table \ref{table_results_benchmark} shows that our \emph{S-MAG} scheme outperforms \emph{SOTA} methods especially considering that authors of \q{\emph{NN strat}} claimed their method outperforms \emph{SOTA} ones. Besides, note that, result of \emph{NN-strat} had been obtained using 1000 iterations, a batch size set to 16384 and it had been shown by its authors that \emph{Adam} algorithm is not suitable. Our multitask modelling allows to reduce the needs of a large training dataset and high number of iterations and, the classic \emph{Adam} optimizer proves to be sufficient. 

\vspace{0.1cm}
Furthermore, in this example, the set $\mathcal{Q}_k$ may be large at some dates yielding several tasks to train. This inclined us to think that basic weight strategies (\emph{UW}, \emph{EW}) as well as Longstaff and Schwartz (LS) method do not perform well when dealing with substantial number of tasks. Indeed, as it will be shown in Table \ref{results_our_example_dim_1_3}, with fewer number of tasks to train together, results across the different methods are comparable. More rigorously, it is well-known that BDPP-based approaches intrinsically incorporate propagation of errors, which in our case, increase as soon as the set $\mathcal{Q}_k$ is large (i.e. there are several tasks). This may explain why, in that cases, BDPP-based approaches like \emph{LS} method are less accurate than global optimization approaches like \emph{NN strat}. Our multitask scheme \emph{S-MAG}, even if based on the BDPP, somehow correct this issue mainly due to knowledge transfer across tasks.

\vspace{0.1cm}
We now assess the robustness of our method by increasing the problem dimension. That is, we consider two models: a one-factor and a three-factor diffusion models. In the remaining we will refer to as \emph{base case setting} the following configurations: $\underline{q} = 0, \overline{q} = 1, \underline{Q} = 20, \overline{Q} = 25, F_{0, t_k} = 20, K = 20$.

\vspace{0.1cm}
\noindent
\textbf{One-factor (\bm{$d = 1$}).} We set $\alpha = 4, \sigma = 0.7$. 

\vspace{0.1cm}
\noindent
\textbf{Three-factor (\bm{$d = 3$}).} We set $\alpha_i = 3, \sigma_i =0.25 , \rho_{i,i}= 1, \rho_{i, j} = 0.3$ for $i \neq j$.

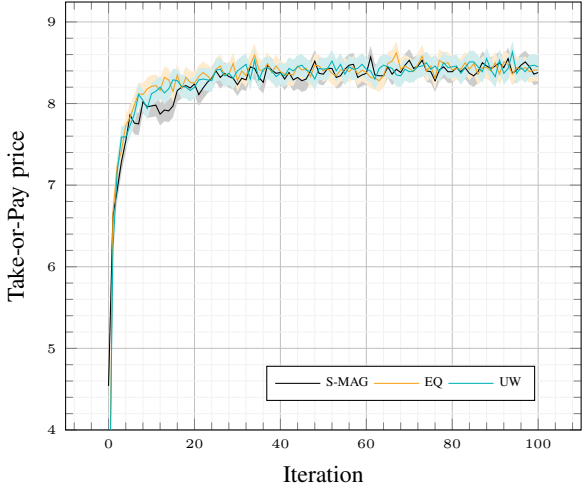
\begin{figure}[httb]
    \begin{tikzpicture}
        \begin{axis}[
            label style={font=\small},
            xlabel={Iteration},
            ylabel={Take-or-Pay price},
            ymin = 4, % remove pour tout voir
            grid=both,
            grid style={line width=.1pt, draw=gray!10},
            major grid style={line width=.2pt,draw=gray!50},
            ticklabel style={font=\tiny},
            axis line style={latex-latex},
            minor tick num=4,
            legend style={at={(0.65,0.15)},anchor=north, font=\tiny},
            legend columns=5
            ]

\addplot[color = black] table[x =iter ,y=price_ema, col sep = comma, mark = none] {loss_log.csv};
\addlegendentry{S-MAG};

\addplot[color = YellowOrange] table[x =iter ,y=price_eq, col sep = comma, mark = none] {loss_log.csv};
\addlegendentry{EQ};

\addplot[color = BlueGreen] table[x =iter ,y=price_unif, col sep = comma, mark = none] {loss_log.csv};
\addlegendentry{UW};

\addplot [name path=upperSmag,draw=none] table[x =iter ,y = price_ema_u, col sep = comma] {loss_log.csv};
\addplot [name path=lowerSmag,draw=none] table[x =iter ,y = price_ema_l, col sep = comma] {loss_log.csv};
\addplot [fill=black!20] fill between[of=upperSmag and lowerSmag];

\addplot [name path=upperEq,draw=none] table[x =iter ,y = price_eq_u, col sep = comma] {loss_log.csv};
\addplot [name path=lowerEq,draw=none] table[x =iter ,y = price_eq_l, col sep = comma] {loss_log.csv};
\addplot [fill=YellowOrange!20] fill between[of=upperEq and lowerEq];

\addplot [name path=upperU,draw=none] table[x =iter ,y = price_unif_u, col sep = comma] {loss_log.csv};
\addplot [name path=lowerU,draw=none] table[x =iter ,y = price_unif_l, col sep = comma] {loss_log.csv};
\addplot [fill=BlueGreen!20] fill between[of=upperU and lowerU];

        \end{axis}
    \end{tikzpicture}
    \caption{Take-or-Pay price estimation in the one-factor model and for the first 100 iterations. The shaded areas represent confidence interval.}
    \label{one_factor_take_or_pay_learning_curve}
\end{figure}

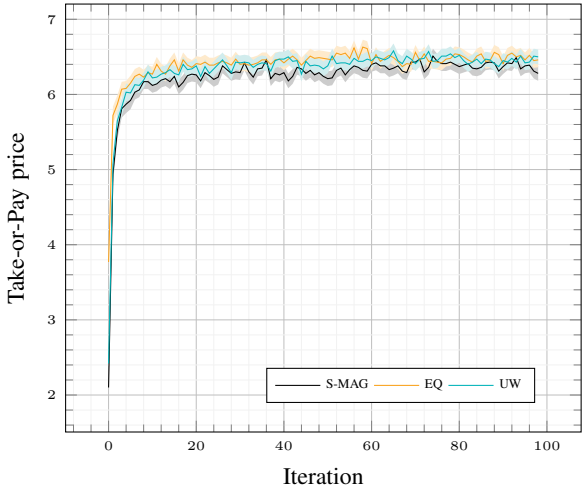
\begin{figure}[httb]
    \begin{tikzpicture}
        \begin{axis}[
            label style={font=\small},
            xlabel={Iteration},
            ylabel={Take-or-Pay price},
            grid=both,
            grid style={line width=.1pt, draw=gray!10},
            major grid style={line width=.2pt,draw=gray!50},
            ticklabel style={font=\tiny},
            axis line style={latex-latex},
            minor tick num=4,
            legend style={at={(0.65,0.15)},anchor=north, font=\tiny},
            legend columns=5
            ]

\addplot[color = black] table[x =iter ,y=price_ema, col sep = comma, mark = none] {loss_log_q3.csv};
\addlegendentry{S-MAG};

\addplot[color = YellowOrange] table[x =iter ,y=price_eq, col sep = comma, mark = none] {loss_log_q3.csv};
\addlegendentry{EQ};

\addplot[color = BlueGreen] table[x =iter ,y=price_unif, col sep = comma, mark = none] {loss_log_q3.csv};
\addlegendentry{UW};

\addplot [name path=upperSmag,draw=none] table[x =iter ,y = price_ema_u, col sep = comma] {loss_log_q3.csv};
\addplot [name path=lowerSmag,draw=none] table[x =iter ,y = price_ema_l, col sep = comma] {loss_log_q3.csv};
\addplot [fill=black!20] fill between[of=upperSmag and lowerSmag];

\addplot [name path=upperEq,draw=none] table[x =iter ,y = price_eq_u, col sep = comma] {loss_log_q3.csv};
\addplot [name path=lowerEq,draw=none] table[x =iter ,y = price_eq_l, col sep = comma] {loss_log_q3.csv};
\addplot [fill=YellowOrange!20] fill between[of=upperEq and lowerEq];

\addplot [name path=upperU,draw=none] table[x =iter ,y = price_unif_u, col sep = comma] {loss_log_q3.csv};
\addplot [name path=lowerU,draw=none] table[x =iter ,y = price_unif_l, col sep = comma] {loss_log_q3.csv};
\addplot [fill=BlueGreen!20] fill between[of=upperU and lowerU];

        \end{axis}
    \end{tikzpicture}
    \caption{Take-or-Pay price estimation in the three-factor model and for the first 100 iterations. The shaded areas represent confidence interval.}
    \label{three_factor_take_or_pay_learning_curve}
\end{figure}

\vspace{0.1cm}
The learning curves for both models in the \emph{base case setting} are depicted in Fig. \ref{one_factor_take_or_pay_learning_curve} and Fig. \ref{three_factor_take_or_pay_learning_curve}. Comparison with benchmark is deferred to Table \ref{results_our_example_dim_1_3} and Table `\ref{table_results_benchmark_robust_comp}.

\begin{table}[htbp]
\caption{Comparison for the pricing of Take-or-Pay contracts with respect to the problem dimension.}
\begin{center}
\begin{tabular}{|c|c|c|c|c|c|}
\hline
& \multicolumn{3}{c|}{Multitask} & \multicolumn{2}{c|}{Benchmark}\\
\hline
& \textbf{S-MAG}& \textbf{EW}& \textbf{UW} & \textbf{NN-strat}& \textbf{LS} \\
\hline
ToP ($d=1$) & 8.45 & 8.41 & \textbf{8.47} & 8.36 & 8.44 \\
\hline
ToP ($d=3$) & 6.47 & \textbf{6.50} & 6.45 & 6.40 & 6.45 \\
\hline
\end{tabular}
\label{results_our_example_dim_1_3}
\end{center}
\end{table}

\vspace{0.1cm}
In Table \ref{table_results_benchmark_robust_comp}, we present a comparison of our method with \emph{SOTA} methods in different settings. Contract $i$ for $i \in \{1,2,3\}$ corresponds to \emph{base case setting} but where we respectively set: $K = 19, \underline{Q} = 15, \underline{Q} = 0$.

\begin{table}[htbp]
\caption{Comparison for the pricing of Take-or-Pay contracts with different settings and different problem dimensions.}
\begin{center}
\begin{tabular}{|c|c|c|c|c|c|c|}
\hline
& & \multicolumn{3}{c|}{Multitask} & \multicolumn{2}{c|}{Benchmark}\\
\hline
& & \textbf{S-MAG}& \textbf{EW}& \textbf{UW} & \textbf{NN-strat} & \textbf{LS} \\
\hline
\multirow{3}{3em}{$d=1$}  & Contract 1 & 31.00 & \textbf{31.01} & 30.99 & 30.98 & 30.98 \\
& Contract 2 & 14.85 & \textbf{14.89} & \textbf{14.89} & 14.84 & 14.80 \\
& Contract 3 & \textbf{28.37} & \textbf{28.37} & 28.31 & 28.36 & 28.34 \\
\hline
\multirow{3}{3em}{$d=3$} & Contract 1 & 29.12 & \textbf{29.15} & 29.09 & 29.12 & 29.10 \\
& Contract 2 & 11.66 & 11.67 & \textbf{11.68} & 11.66 & 11.65\\
& Contract 3 & \textbf{22.72} & 22.70 & 22.65 & 22.71 & 22.68 \\
\hline
\end{tabular}
\label{table_results_benchmark_robust_comp}
\end{center}
\end{table}

\noindent
From Table \ref{results_our_example_dim_1_3} and Table \ref{table_results_benchmark_robust_comp}, it appears that our multitask modelling consistently performs well regardless of the problem dimension. Furthermore, as previously asserted, when there are fewer number of tasks, all methods are comparable. As proved in Table \ref{table_results_benchmark}, our scheme \emph{S-MAG} stands out as soon as the problem becomes more complex i.e. with large set $\mathcal{Q}_k$ meaning several tasks to train. In the latter case, unlike our multitask modelling, the other BDPP-based methods like \emph{LS} method become ineffective. Besides, it is important to notice that in our experiments, it appeared that when the contract is \emph{in-the-money} i.e. $S_0 > K$, all methods are also comparable. But, once again, our multitask modelling \emph{S-MAG} stands out when the contract is \emph{at-the-money} i.e. $S_0 = K$; where the optimal control is more difficult to find.

\vspace{0.1cm}
\noindent
\textbf{Network architecture \emph{Versus} weights scheme.} The choice of the multitask network architecture and the weight scheme is pivotal and, at times, complementary. In certain problems \cite{b15}, having a \q{good} network architecture is sufficient, and basic weight strategies prove effective. Conversely, there are situations \cite{b5} where the choice of the loss weight strategy outweighs the importance of the network architecture. In some cases, both the network architecture and the weight strategy play crucial roles. It appears that our Take-or-Pay contract pricing problem falls into the latter category. Indeed, our experiments revealed that omitting the introduction of the feature $m(Q)$ (refer to \eqref{def_decision_function_f}) in our network resulted in poor performance. Furthermore, basic weight strategies consistently failed to produce satisfactory results in all instances specially when training several tasks together.

\subsection{Swing contract with penalty}
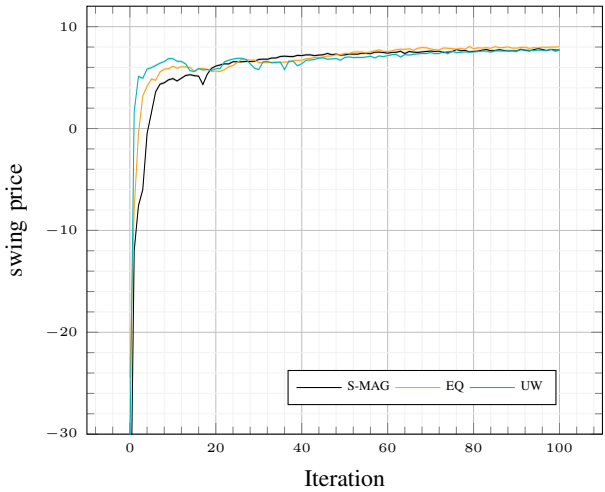
\begin{figure}[httb]
    \begin{tikzpicture}
        \begin{axis}[
            label style={font=\small},
            xlabel={Iteration},
            ylabel={swing price},
            ymin = -30,
            grid=both,
            grid style={line width=.1pt, draw=gray!10},
            major grid style={line width=.2pt,draw=gray!50},
            ticklabel style={font=\tiny},
            axis line style={latex-latex},
            minor tick num=4,
            legend style={at={(0.65,0.15)},anchor=north, font=\tiny},
            legend columns=5
            ]

\addplot[color = black] table[x =iter ,y=price_ema, col sep = comma, mark = none] {loss_log_penalty_2e_test.csv};
\addlegendentry{S-MAG};

\addplot[color = YellowOrange] table[x =iter ,y=price_eq, col sep = comma, mark = none] {loss_log_penalty_2e_test.csv};
\addlegendentry{EQ};

\addplot[color = BlueGreen] table[x =iter ,y=price_unif, col sep = comma, mark = none] {loss_log_penalty_2e_test.csv};
\addlegendentry{UW};

\addplot [name path=upperSmag,draw=none] table[x =iter ,y = price_ema_u, col sep = comma] {loss_log_penalty_2e_test.csv};
\addplot [name path=lowerSmag,draw=none] table[x =iter ,y = price_ema_l, col sep = comma] {loss_log_penalty_2e_test.csv};
\addplot [fill=black!20] fill between[of=upperSmag and lowerSmag];

\addplot [name path=upperEq,draw=none] table[x =iter ,y = price_eq_u, col sep = comma] {loss_log_penalty_2e_test.csv};
\addplot [name path=lowerEq,draw=none] table[x =iter ,y = price_eq_l, col sep = comma] {loss_log_penalty_2e_test.csv};
\addplot [fill=YellowOrange!20] fill between[of=upperEq and lowerEq];

\addplot [name path=upperU,draw=none] table[x =iter ,y = price_unif_u, col sep = comma] {loss_log_penalty_2e_test.csv};
\addplot [name path=lowerU,draw=none] table[x =iter ,y = price_unif_l, col sep = comma] {loss_log_penalty_2e_test.csv};
\addplot [fill=BlueGreen!20] fill between[of=upperU and lowerU];

        \end{axis}
    \end{tikzpicture}
    \caption{Swing (with penalty) price estimation for the one-factor model and for the first 100 iterations. The shaded areas represent confidence interval.}
    \label{learning_curve_with_pen}
\end{figure}

\noindent
In the swing contract with penalty \cite{b1, b3}, the holder of the contract wants to solve \eqref{soc_pb} with
\begin{equation}
    \label{def_soc_pb_esp_swing_penalty}
    g_n(s, Q) = -S\big(A(Q - Q_A)_{-} + B(Q-Q_B)_{+}\big),
\end{equation}
\noindent
where $A, B$ are positive real constants. The function $c_k$ is the same as in \eqref{def_soc_pb_esp_swing}. In this case, we consider a $n=30$-dates swing contract. We set $\underline{q} = 0, \overline{q} = 1, Q_A = 20, Q_B = 25, F_{0, t_k} = 20, K = 20, A = B = 1$. As already mentioned, in this contract the holder may violate global constraints but not local ones. Thus, in this case, we have $\mathcal{Q}_k := \mathbb{N} \cap [k_n \underline{q}, k_n \overline{q}]$ with $k_n := \min(k, n-1)$. Learning curves are depicted in Fig \ref{learning_curve_with_pen}.

\section{Conclusion}
\noindent
We have introduced a highly efficient and effective method using multitask neural networks to address some stochastic optimal control problems that classic neural networks may struggle to solve efficiently. To train our multitask neural networks, we also introduced a novel loss weighting scheme. To prove the applicability of our multitask modelling, we focused on the pricing of swing option, a commodity derivative product. Notably, this paper pioneers the use of neural networks for pricing such financial products, based on the backward dynamic programming equation. We proved that our multitask approach is robust and outperforms state-of-the-art methods.

\section*{Acknowledgment}
\noindent
The first author would like to thank Asma Meziou, Gilles Pagès and Vincent Lemaire for fruitful discussions and comments.

\end{document}